\documentclass{svproc}
\usepackage{amsmath,amssymb,amsfonts,bm}
\usepackage{algorithmic}
\usepackage{graphicx}
\usepackage{cite}
\usepackage{mathtools}
\usepackage{subcaption}
\usepackage{cancel}
\usepackage[normalem]{ulem}
\usepackage[ruled,linesnumbered,noend]{algorithm2e}
\usepackage[colorlinks=true, allcolors=blue]{hyperref}

\newcommand{\vb}[1]{\textbf{#1}}
\newcommand{\norm}[1]{\left\lVert#1\right\rVert}

\newcommand{\mt}[1]{\mathrm{#1}}
\newtheorem{assumption}{Assumption}

\usepackage{url}

\begin{document}
\mainmatter              
\title{Multi-Source Encapsulation With Guaranteed Convergence Using Minimalist Robots}
\titlerunning{Multi-Source Encapsulation}  
%
\author{Himani Sinhmar \and Hadas Kress-Gazit}
\authorrunning{Himani Sinhmar, Hadas Kress-Gazit} 
%
\tocauthor{Himani Sinhmar, Hadas Kress-Gazit}
\institute{Cornell University, Ithaca NY 14853, USA,\\
\email{hs962@cornell.edu}, 
\email{hkg@cornell.edu}}

\maketitle              

\begin{abstract}
We present a decentralized control algorithm for a minimalist robotic swarm lacking memory, explicit communication, or relative position information, to encapsulate multiple diffusive target sources in a bounded environment. The state-of-the-art approaches generally require either local communication or relative localization to provide guarantees of convergence and safety. We quantify trade-offs between task, control, and robot parameters for guaranteed safe convergence to all the sources. Furthermore, our algorithm is robust to occlusions and noise in the sensor measurements as we demonstrate in simulation.
\keywords{Minimalist robots, Swarms, Multi-Source encapsulation, Obstacle avoidance, Guaranteed convergence}
\end{abstract}

\setlength{\textfloatsep}{0.2pt}
\section{Introduction}
Signal source detection, localization, and encapsulation play a critical role across biological systems, spanning from chemotaxis in microscopic organisms~\cite{berg1975chemotaxis} to predators' olfactory tracking~\cite{hughes2010predators}. In robotics, researchers develop sensor equipped robots for identifying signal sources for applications such as  nanomedicine~\cite{li2017nanomedicineReview}, chemical hazard detection~\cite{ishida2012hazardusReview}, nuclear radiation tracing~\cite{nagatani2013nuclearFukushima}, and environmental monitoring~\cite{sung2019envMoni}. Extensive research exists for source localization using either individual agents or a network of agents. A common approach involves gradient computation, either directly through sensors capable of measuring gradients~\cite{MAITI} or by robots that deduce the gradient by sequentially measuring the field's concentration and making incremental movements. Using multiple robots as a network can enhance accuracy and efficiency by sharing information for collective gradient estimation at each time-step~\cite{mahdi2012adaControl}, requiring less memory and computational power. However, this method depends on inter-robot communication and the ability to discern their neighbors' signal direction.

In this paper, we consider multi-source encapsulation using a swarm of minimalist robots (no memory, no self-localization, no direct communication, and inability to determine neighbors' relative locations) equipped only with omnidirectional sensors~\cite{sinhmar2022decentralized} to detect the concentration of environmental signals. Here, each robot independently estimates signal gradients at each time step without requiring information from its neighbors or memory for storing past concentration values. This method is particularly useful in contexts where traditional navigation and communication technologies are unfeasible, such as nano-medicine for tumor encapsulation~\cite{li2017nanomedicineReview} or in underwater environments~\cite{underwater_2} where restricted communication bandwidth challenges the coordination of autonomous vehicles. Previously~\cite{sinhmar2022decentralized,sinhmar2023guaranteed}, we established the mathematical foundation for a swarm of minimalist robots to encapsulate multiple targets with non-overlapping influence regions. In this paper, we tackle multi-source encapsulation with overlapping influence regions. A target is encapsulated if a user-specified number of robots are present simultaneously in a user-specified encapsulation ring around it. 

In contrast to source localization where the robots can follow a gradient, encapsulation is more complex since it also involves accurately determining the encirclement contour for each target. By treating each target as a point source with no encapsulation ring, the multi-source encapsulation problem boils down to the localization of all peaks of a spatiotemporal multi-modal function.

\noindent\textbf{Related Work:}
While the literature covers the detection of a single source by its gradient extensively~\cite{du2021single,li2012single,li2020single,lin2017single,azuma2012single,drone2022single,azuma2015single,ghods2011single,fluids2009single,hyprid2008single,casbeer2005single,atanasov2015single,sukhatme2004single,distibuted2016single,distibuted2018single}, the simultaneous detection of multiple overlapping sources is not as well explored~\cite{mcgill2009single}. Most existing source-seeking strategies either require communication between the agents and/or relative position information of an agent's neighbors. In~\cite{ghosh2006glowworm,ghosh2008theoretical,ghosh2009glowworm}, the authors introduced a glowworm swarm optimization algorithm (GSO) designed for multiple source localization. Each robot can determine its position relative to others and avoid collisions. They share signal information to select a leader robot, which guides the others. The GSO algorithm cannot guarantee the complete localization of all sources and its effectiveness depends on the initial distribution of the swarm. 
\cite{wu2015SUSD} introduces a fish-school-inspired~\cite{couzin2013fish} distributed approach for single source seeking using mobile sensor networks without the need for agents to estimate gradients. However, for groups larger than two agents, communication among the agents becomes necessary. 
In~\cite{wu2023multi}, authors developed an algorithm for a robotic swarm to encapsulate diffusive targets among convex obstacles by using relative position information to build an artificial gene network for coordinated movement toward targets and used simulations for verification of the algorithm. In~\cite{sukhatme2007detecting}, authors utilize communication between a mobile sensor and uniformly distributed localized static sensor nodes to perform gradient descent for contour following. \cite{kitts2018adaptive} introduces a reactive control policy for holonomic robot clusters for tasks like extrema finding, contour following, ridge/trench tracking, and saddle point keeping. This strategy relies on the cluster maintaining a fixed geometry optimized for scalar field navigation effectively turning the cluster into a larger entity with distributed sensors along its perimeter.

There is less work that provides guarantees of convergence for source encapsulation using minimalist robots. One of the earliest efforts to establish a theoretical basis for source localization is in~\cite{ogren2004naomi,moreau2003naomi,leonard2001naomi,bachmayer2002naomi} using projected gradients and artificial potential functions to maintain formation, hence requiring all-to-all communication. In~\cite{robinett2004stability}, the stability of a system, wherein each robot estimates a second-order representation of an unknown scalar field through a communication network, is validated using Lyapunov analysis. In contrast, we do not estimate the scalar field, leading to a control policy that requires less computational effort.
There also exists work on using maximum likelihood localization~\cite{baidoo2016mle},  optimization~\cite{du2022opti,lee2019opti,zhang2011opti,mishra2016opti,ramirez2018opti,multiple2018opti,sakurama2016opti,gronemeyer2017opti,li2021bayesianLearningOpti}, and machine learning techniques~\cite{al2018learning,li2023learning,duisterhof2019learning,li2021bayesianLearningOpti} for multiple source localization. These approaches generally diverge from the use of minimalist robots, which is central to our work.

\noindent\textbf{Contribution}: The novelty of this paper is threefold: (i) we introduce a scalable, decentralized control law for a minimalist robotic swarm (each robot is devoid of memory, self-localization, and explicit communication ability) that can encapsulate multiple signal sources (targets) without the need for accurate detection of relative positions of the targets, neighboring robots, or obstacles;  (ii) we present a derivative-free algorithm based on the simplex gradient to identify multiple optima of a multimodal scalar spatiotemporal field in an obstacle-cluttered environment using distributed sensing; (iii) we offer theoretical guarantees for convergence and provide bounds on task, control, and robot parameters to ensure all targets are encapsulated safely.
Furthermore, we analyze the robustness of the control algorithm to sensor noise, occlusions, and asynchronous execution. 

\section{Environment and Robot Model}
\label{section_prelim}
We adopt the environment, robot, and target models from~\cite{sinhmar2023guaranteed}, adding convex obstacles in this work. We consider a $2D$ bounded \textbf{environment} $E \subseteq \mathbb{R}^2$ with a fixed global frame {$\mathcal{I}$}. A \textbf{target} $g =(\mathbf{c}_{g},r_{g})$ is a disk of radius $r_{g} \geq 0$ centered at $\mathbf{c}_{g}\in \mathcal{I}$. The set $O = \{O_1 \cdots O_w\}$ represents closed convex obstacles in the environment. A \textbf{robot}, $R = (\mathbf{c}_r,\gamma_r,r_{r},p,Z)$, is a disk of radius $r_{r} > 0$ centered at $\mathbf{c}_r \in \mathcal{I}$ with heading $\gamma_r \in \mathbb{S}$.  The following set of equations gives the differential drive kinematics of a robot: $\gamma_{r,T} = \gamma_{r,T-1} + \theta_r$ and $\vb{c}_{r,T} = \vb{c}_{r,T-1} + d_r[\mt{cos}\gamma_{r,T} \quad \mt{sin}\gamma_{r,T}]$. At each time step T, the robot is controlled in a \textit{turn-then-move} scheme with control inputs $\theta_r \in \mathbb{S}$ and $d_r \in \mathbb{R}^+$ with the step-size upper bounded as $d_r \leq d_r^{\mt{max}}$. Robots are reactive, memoryless, with no knowledge of others' locations or communication ability, and equipped with $p$ isotropic sensors around their periphery. Angles $\phi^k$ for $k \in \{1 \cdots p\}$ denote each sensor's orientation relative to the robot's heading. Sensor measurements are aggregated in the set $Z$. The robot's frame, $\mathcal{B}$, is centered at $\vb{c}_r$, with its $x-$axis always aligned with the heading direction $\gamma_r$.

We consider four types of signal-emitting sources detectable by a robot's sensor: point sources located at the center of each target ($s_g$), each robot ($s_r$), each obstacle ($s_o$), and a line source spanning the boundary of the environment ($s_e$). For simplicity, we refer to these signal types as $\{g, r, o, e\}$ with a signal strength function $f_s: \mathbb{R}^2 \rightarrow \mathbb{R}$. The influence radius of a source is limited to $\beta_s$, such that at a point $\vb{x} \in \mathcal{I}$ and a source located at $\vb{S} \in \mathcal{I}$, $f_s(\vb{x}) = 0 \textrm{ } \mt{if} \norm{\vb{x}-\vb{S}} \geq \beta_s$. For brevity, we define $d=\norm{\vb{x}-\vb{S}}$. 
For the $k^{\text{th}}$ sensor, $N^k_s$ denotes the set of sources of type $s$ within its sensing range, with $d_j^k$ representing the distance to source $j \in N^k_s$. The sensor's reading, $z_s^k$, is the sum of signal strengths from all $N^k_s$ sources, $z_s^k = \sum_{j \in N^k_s} f_s(d^{k}_j)$. For a line source within $\beta_e$, this sum transforms into an integral along the boundary segment. The tuple $(z_g^k,z_r^k,z_o^k,z_e^k)$ corresponds to the measurements of the $k^{th}$ sensor and $Z=\cup_k (z_g^k,z_r^k,z_o^k,z_e^k)$. We define user-specified safe distances $r^{\text{safe}}_s \textrm{ }, s \in \{g,r,o,e\}$ as the minimum distance a robot is required to keep from a source. 
An encapsulation ring is an annular region with an inner radius $r^\mt{safe}_g$ and outer radius $r^\mt{encap}_g$ centered at $\mathbf{c}_{g}$.  A robot is considered to be in the encapsulation ring if, $r^\mt{safe}_g<\norm{\vb{c}_{r,T}-\vb{c}_{g}}\leq r^\mt{encap}_g$.  A target is considered \textit{encapsulated} when $n_g$ robots, a user-specified number, occupy its encapsulation ring concurrently.

\section{Problem Formulation}
\label{section_Problem}
We consider $n$ robots each equipped with $p$ sensors, $m$ static targets, and $w$ static obstacles in a bounded environment $E$ with a random initial distribution and no information about their location and the locations of targets, obstacles, and other robots. Given safe distances $r^{\mt{safe}}_s, s \in \{g,r,o,e\}$, and the encapsulation ring parameters ($r^\mt{encap}_g$, $n_g$) such that $\sum_{g \in \mathcal{G}} n_g \leq n$ we find a real-time decentralized control law for each robot such that all targets are eventually encapsulated while maintaining minimum safety distances at all times.
We make the following assumptions:
\begin{assumption}\label{assum_sensorPlacement}
    The sensor placement is such that when a robot's center is $r_s^{\mt{safe}}$ away from a source $s$, at least one sensor is in the influence region of the source.
\end{assumption}
\begin{assumption}\label{assum_targetDistance}
    We constrain a target to maintain a minimum distance of ($r_g^{\mt{encap}}+ r_e^{\mt{safe}} + d_r^{\mt{max}}$) from the environment boundary. This ensures that robots can encapsulate the target without colliding with the environment boundary. 
\end{assumption}
\begin{assumption}\label{assum_targetShutOff}
    On encapsulation, a target emits a single burst of a shut-off signal and stops emitting any signal after that. Robots within the encapsulation ring stop moving ($d_r=0$) and start emitting an obstacle-type signal $s_o$.
\end{assumption}
\begin{assumption}\label{assum_fknown}
    The function $f_s(d)$ is convex and decreases monotonically as the distance $d$ from a source increases. It is continuously differentiable, with its gradient being L-Lipschitz continuous. For a point $P$ at position $\vb{x} \in \mathcal{I}$ influenced by $m$ sources of type $s$ at positions $\vb{S}_1, \vb{S}_2, \ldots, \vb{S}_m \in \mathcal{I}$, the aggregate signal strength at $P$ is defined by $F_s(\vb{x}) = \sum_{i=1}^{m}f_s(\norm{\vb{x}-\vb{S}_i})$. The function $F_s$ (sum of convex functions) adheres to the continuity and differentiable properties of its constituent functions~\protect{\cite[Chapter 3]{boyd2004convex}}. Furthermore, the robots know the function $f_s(d)$ and its inverse $\forall s \in \{g,r,o,e\}$. However, it does not know the aggregate signal function $F_s$, hence it lacks comprehensive information about the combined effects of multiple sources.
\end{assumption}
\begin{assumption}
    There is no static obstacle in the encapsulation ring of a target.
\end{assumption}

\subsection{Mathematical Preliminaries}
\label{subsection_math}
To direct a robot toward the closest target without knowing the number or positions of targets, we calculate gradients using its sensor readings. The simplex gradient method approximates gradients where direct derivative calculation is unfeasible, by constructing an affine approximation of the function near a point of interest. It selects points within the function's domain to form a simplex, and the gradient is derived from the original function's evaluations at these points.
\begin{lemma} \label{def_simplex}
    \textbf{Simplex Gradient}:\cite{regis2015calculus} Let $F:\mathbb{R}^n \rightarrow \mathbb{R}$ and $\mathcal{X} = \{\vb{x}_0,\vb{x}_1 \cdots \vb{x}_k\}$ be an ordered set of $k+1$ affine independent points in $\mathbb{R}^n$ such that $k \geq n$. Define, $H(\mathcal{X}) = [\vb{x}_1-\vb{x}_0 \cdots \vb{x}_k-\vb{x}_0]\in \mathbb{R}^{n \times k}$ and $\delta_F(\mathcal{X}) = [F(\vb{x}_1) - F(\vb{x}_0) \cdots F(\vb{x}_k) - F(\vb{x}_0)]^T
    \in \mathbb{R}^k$. Then for $k=n$, $H(\mathcal{X})$ is invertible and the simplex gradient of $F$ at $\vb{x}_0$ over $H(\mathcal{X})$ is given by $\nabla_{\mt{\mt{SG}}} F(\vb{x}_0;H) = H^{-\mt{T}}\delta_F(\mathcal{X})$.
\end{lemma}

\begin{lemma}\label{def_errSimplex}
    \textbf{Error bound for Simplex Gradient}:\cite{regis2015calculus} Let $F$ be a continuously differentiable function in an open domain $\Omega$ containing the closed ball $B(\vb{x}_0,\Delta_H) = \{\vb{x} \mid \norm{\vb{x}-\vb{x}_0} \leq \Delta_H\}$, where $\Delta_H = \mt{max}_{1 \leq i\leq k}\norm{\vb{x}_i-\vb{x}_0}$. Further, assume that $\nabla F$ is  Lipschitz continuous in $\Omega$ with Lipschitz constant $L \geq 0$ and
    $\Hat{H}(\mathcal{X})^{\mt{T}} = H(\mathcal{X})^{\mt{T}}/\Delta$. Then, $\norm{\nabla_{\mt{\mt{SG}}}F(\vb{x}_0) - \nabla F(\vb{x}_0)} \leq k^{1/2}\frac{L}{2}\Delta_H$.
\end{lemma}

\section{Approach}
In our prior work~\cite{sinhmar2022decentralized,sinhmar2023guaranteed}, we considered targets with non-overlapping influence regions, i.e. a robot can only sense one target at a time. A robot then estimates the position of the sensed target using the sensor receiving the strongest signal. When considering overlapping influence regions, two challenges emerge due to aggregated signals: first, the sensor with the strongest signal no longer reliably indicates the direction to the closest target; second, it is challenging to compute the distance to a target. In this paper, we develop a control policy based on fusing distributed information from a robot's sensors to encapsulate all targets while maintaining a safety distance from obstacles, other robots, and targets. In Section~\ref{section_ca}, we derive bounds on $d_r$ and $\beta_r$ so that a robot maintains user-specified safety distances. We relax the upper bound, with respect to~\cite{sinhmar2023guaranteed}, on the step size a robot can take.
In Section~\ref{section_ta}, we utilize simplex gradients to estimate a robot's heading for motion toward the closest target. In Section~\ref{section_control}, we present the reactive control policy for a robot to encapsulate a target.

\subsection{Maintaining safety distances}\label{section_ca}
Given a robot is equipped with isotropic sensors, the same measurement could correspond to a single source nearby or a cluster of sources further away. Therefore, for each sensor reading, $z_s^k, s \in \{r,o,e\}$, we define a virtual source on a circle centered at the sensor $k$ such that the closest possible location of the virtual source with respect to the robot's center is given by, 
\begin{equation}\label{dsk}
d_s = r_r\mt{cos}(\pi/p) + \sqrt{(d_s^{k})^2-r_r^2\mt{sin}^2(\pi/p)}    
\end{equation}
Using Assumption~\ref{assum_fknown}, we can say that there cannot be an actual source closer to the robot than the virtual source~\cite{sinhmar2022decentralized}. For asymmetric sensor placement, $\pi/p$ is substituted with half the largest angle formed by the $k^{\text{th}}$ sensor and either of its neighboring sensors. Likewise, for robots with non-circular geometries, $r_r$ is adjusted to reflect the distance from the $k^{\text{th}}$ sensor to the robot's center.

At each time step, the robot estimates the relative distance between its center and the nearby sources using Eq.~\eqref{dsk} for the sensor with the maximum sensor reading $z_s^k = \mt{max}(Z_s)$. If this distance is less than or equal to $(r_s^{\mt{safe}} + d_s^{\mt{max}})$, where $d_s^{\mt{max}}$ is the maximum step-size of the source $s$, a collision avoidance behavior is triggered.
\begin{figure}[!b]
  \centering
  \includegraphics[width=0.8\linewidth]{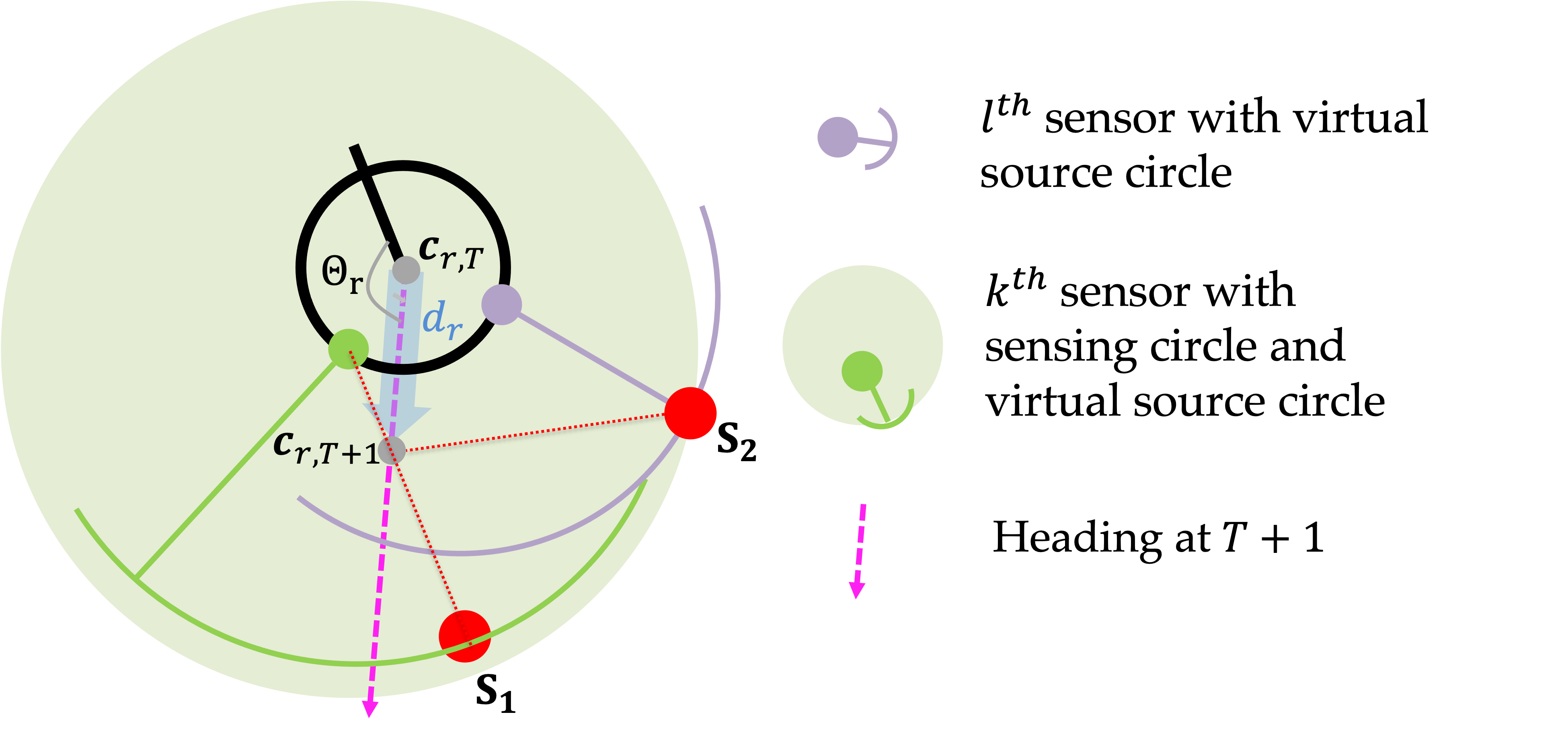}
  \caption{Computing $d_r$ that ensures a safe distance from nearby obstacles}
  \label{fig:dynColl}
\end{figure}
The step-size $d_{r}$ that a robot takes at time step $T$ in a given heading direction $\gamma_{r, T+1}$ must be chosen such that, at next time step $T+1$, the robot maintains at least a distance of $r_s^{\mt{safe}}$ from the closest virtual source of type $s$. As shown in Fig.~\ref{fig:dynColl}, let $k$ and $l$ be the indices of the sensors closest to the intended heading direction $\gamma_{r, T+1}$ with $d_s^k$ and $d_s^{\ell}$ being the radii of virtual sources respectively such that $d_s^k > d_s^{\ell}$.  If we consider only the sensor reading from the $k^{\mt{th}}$ sensor, the closest virtual source to the robot is located at $\vb{S}_1 \in \mathcal{I}$ as depicted in Fig.~\ref{fig:dynColl}. Then the step-size $d_{r}$ that the robot can take should be such that $\norm{\vb{c}_{r,T+1}-\vb{S}_1} \geq r_s^{\mt{safe}} + d_s^{\mt{max}}$ as given by Eq.~\eqref{avoObsGivenTh_S1}.
\begin{equation}
    \label{avoObsGivenTh_S1}
    0 \leq d_{r} \leq r_r \mt{cos}(\phi^{k}-\theta_r) + \sqrt{(d_r^{k} - r_r^{\mt{safe}} - d_s^{\mt{max}})^2 - r_r^2 \mt{sin}^2(\phi^{k}-\theta_r)}
\end{equation}
Sources that lie outside the sensing radius of the $k^{\mt{th}}$ sensor (shaded green region in Fig.~\ref{fig:dynColl}) may create a virtual source that is closer than the one described above. Therefore, the closest virtual source could also be at the intersection of the virtual source circle of the $\ell^{\mt{th}}$ sensor and the sensing circle of the $k^{\mt{th}}$ sensor at $\vb{S}_2 \in \mathcal{I}$. The robot should maintain a safe distance from $\vb{S}_2$ as well, i.e. $\norm{\vb{c}_{r,T+1}-\vb{S}_2} \geq r_s^{\mt{safe}} + d_s^{\mt{max}}$. Consider the triangle $\triangle \vb{S}_2\vb{c}_{r,T}\vb{c}_{r,T+1}$, let $b_0 = \norm{\vb{c}_{r,T}-\vb{S}_2}$ 
and $\psi = \angle \vb{S}_2\vb{c}_{r,T}\vb{c}_{r,T+1}$. Then,
\begin{multline}
    \label{avoObsGivenTh_S2}
    d_{r} \leq b_0 \mt{cos}(\psi) - \sqrt{(r_r^{\mt{safe}} + d_s^{\mt{max}})^2 - b_0^2 \mt{sin}^2(\psi)} \\ \mt{OR}\quad d_r \geq \;
    b_0 \mt{cos}(\psi) + \sqrt{(r_r^{\mt{safe}} + d_s^{\mt{max}})^2 - b_0^2 \mt{sin}^2(\psi)}
\end{multline} 
Furthermore, to ensure that two robots never deadlock, the bounds on $d_r^{\mt{max}}$, and the influence region of a robot's source $\beta_r$, is given by Eq.~\eqref{bound_d_r_max} and Eq.~\eqref{bound_beta_r}, respectively. The proof is detailed in Lemma V.3 of~\cite{sinhmar2023guaranteed}.
\begin{equation}
    \label{bound_d_r_max}
    d_r^{\mt{max}} < \frac{r_r^{\mt{safe}}+r_r\mt{cos}(\pi/p)}{2}
    -\frac{\sqrt{(r_r^\mt{safe})^2 + r_r^2 - 2r_rr_r^\mt{safe}\mt{cos}(\pi/p)}}{2}
\end{equation}
\begin{equation}
    \label{bound_beta_r}
    \sqrt{(r_r^\mt{safe})^2 + r_r^2 - 2r_rr_r^\mt{safe}\mt{cos}(\pi/p)} + 2d_r^{\mt{max}} < \beta_r 
    < r_r^{\mt{safe}}+r_r\mt{cos}(\pi/p)
\end{equation}
Given a heading direction $\gamma_{r,T}$, at each time step we choose the maximum possible step size $d_r$ such that it satisfies the bounds of Eq.~\ref{avoObsGivenTh_S1}, Eq.~\ref{avoObsGivenTh_S2}, and Eq.~\ref{bound_d_r_max} for a source $s \in \{r,o,e\}$. To maintain a safe distance from the environment boundary, the robot's heading direction $\theta_r$ must be chosen from the angular range $\Theta_e^{\mt{avo}} = [\phi^k + \pi/p + \pi/2, \textrm{ } \phi^k - \pi/p + 3\pi/2]$ as derived in~\cite{sinhmar2022decentralized}.

\subsection{Target attraction}\label{section_ta}
The primary objective of a robot's control policy is to navigate toward the closest target by calculating the appropriate heading ($\theta_r$) at every time step and taking a step-size ($d_r$) that ensures safety. The collision avoidance strategy above dictates the step-size. To determine the heading that ensures navigation towards the closest target, we use simplex gradients. Once the robot arrives at the encapsulation ring, it switches to an encirclement behavior. The control policy requires determining the heading direction to move towards a target, and identifying when the robot has entered an encapsulation ring.

\noindent\textbf{\textit{Direction} Towards the Closest Target}: We use a simplex gradient approach (Lemma~\ref{def_simplex}) to estimate the Line of Sight (LOS) vector, $\zeta$, to the closest target. For a given sensor $\ell$ located at $\vb{x}^{\ell} \in \mathcal{B}$, let $H(\mathcal{X}) = [\vb{x}^{\ell-1} - \vb{x}^{\ell} \quad \vb{x}^{\ell+1} - \vb{x}^{\ell}]$ and $\delta{F_g} (\mathcal{X}) = [z_g^{\ell-1} - z_g^{\ell} \quad z_g^{\ell+1} - z_g^{\ell}]^T$, then the simplex gradient at $\vb{x}^{\ell}$ is computed as $\nabla_{\mt{SG}} F_{g}(\vb{x}^{\ell}) = H^{-T}\delta{F_g}(\mathcal{X})$. At each time step, a robot chooses the heading direction dictated by the sensor with the maximum target signal reading. \\
\noindent\textbf{Detecting the \textit{Encapsulation Ring}}: 
When targets' influence regions overlap, a robot's sensor reading cannot be used to estimate the distance to a target. 
Instead, we leverage a key observation: Assumption~\ref{assum_fknown} suggests that the unique signature of the aggregate signal ($F_g$) at any point $\vb{x} \in \mathcal{I}$ can be identified by analyzing the magnitude of the aggregate signal gradient ($\nabla F_g(\vb{x})$) and the expected gradient of a single target's signal ($\nabla f_g(\vb{x})$). For instance, when two targets lie in opposite directions with respect to a sensor, the overall signal strength may be high, but the combined signal gradient's magnitude will be lower than if the signal came from one target only.

\begin{figure}[!b]
  \centering
  \includegraphics[width=0.7\linewidth]{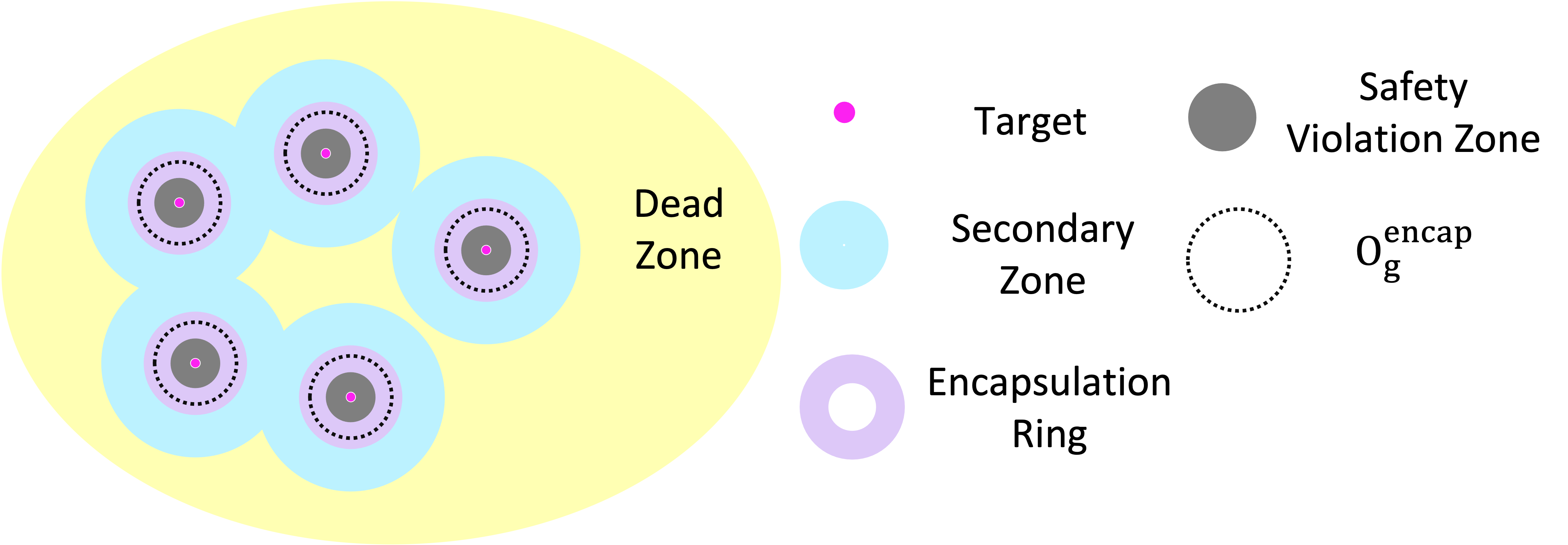}
  \caption{Environment zones.}
  \label{fig:zones}
\end{figure}

Let the $k^{\mt{th}}$ sensor located at $\vb{x}^k \in \mathcal{B}$ be such that $z_g^k = \mt{max}(Z_g)$ and $d_g$ be the robot center's distance to the closest target. 
As shown in Fig.~\ref{fig:zones}, we define different zones that a robot can be in: \textit{secondary zone} if $d_g > r_g^{\mt{encap}}$,  \textit{encapsulation ring} as defined in Section~\ref{section_prelim}, \textit{encirclement contour} $O_g^{\mt{encap}}$, \textit{safety violation zone} if $d_g < r_g^{\mt{safe}}$, and \textit{dead zone} if $z_g^k = 0$. 

For encapsulation, a robot does not need to know the exact distance to the closest target; it only needs to know if it has entered the encapsulation ring to initiate the encirclement behavior. We use the known encapsulation ring parameters and function $f_g(d)$. 
The expected gradient magnitude on the encapsulation ring, assuming the robot's $k^{\mt{th}}$ sensor perceives a single target (i.e. if $F_g = f_g$,) is $\norm{\nabla f_g(r_g^{\mt{encap}})}$. A discrepancy between $\norm{\nabla_{\mt{SG}} F_g(\vb{x}^k)}$ and $\norm{\nabla f_g(r_g^{\mt{encap}})}$ indicates that the robot is not in the encapsulation ring. 

Specifically, $\norm{\nabla_{\mt{SG}} F_g(\vb{x}^k)} < \norm{\nabla f_g(r_g^{\mt{encap}})}$ suggests that the robot is in the secondary zone, whereas $ \norm{\nabla f_g(r_g^{\mt{encap}})} \leq \norm{\nabla_{\mt{SG}} F_g(\vb{x}^k)} < \norm{\nabla f_g(r_g^{\mt{safe}})}$ implies the robot is within the encapsulation ring.  For this methodology to be valid, it is necessary that $F_g \approx f_g$ within a target's encapsulation ring. In Section~\ref{section_guarantees}, we derive constraints on target separation to maintain this condition. Note that a robot can only determine whether it is inside or outside the encapsulation ring; it can neither asse.ss its distance from the encapsulation ring nor from the target.

\subsection{Local Control Law}\label{section_control}
The reactive control policy for a robot is as follows: It first estimates its distance to the environment boundary (Eq.~\eqref{dsk}), if about to violate the safety distance $r_e^{\mt{safe}}$, the heading direction is chosen such that the robot moves away from the boundary ($\theta_r \in \Theta_e^{\mt{avo}}$) with a maximum possible step size $d_r$ while avoiding nearby robots and obstacles as described in Section~\ref{section_ca}. Once the robot is at a safe distance from the boundary, it computes its control parameters depending on which zone it is situated in (Fig.~\ref{fig:zones}). When the robot is present in the dead zone (i.e. $\mt{max}(Z_g) = 0$), it performs a biased random walk, preferring directions dictated by the sensor receiving minimum signal reading from nearby robots. 

\textbf{Secondary zone} ( $\norm{\nabla_{\mt{SG}}F_g(\vb{x}^k)} < \norm{\nabla f_g(r_g^{\mt{encap}})}$): the robot's heading direction is $\theta_r = \zeta$, where $\zeta$ is the LOS vector to the target. If the robot cannot move towards the target due to safety constraints, it moves tangentially to the LOS vector ($\theta_r = \zeta + \pi/2$ or $\theta_r = \zeta + 3\pi/2$ ) preferring the one that minimizes deviation from its current heading. The step-size $d_r$ is chosen to ensure safety with respect to other robots (Section~\ref{section_ca}) and obstacles (Section~\ref{section_obsSecZone}). If the robot can neither move towards a target nor tangentially, it chooses a direction of motion that maximizes the possible step size $d_r$. This behavior prevents temporary deadlocks in the swarm due to local minima created by other dynamic robots between the target and the robot.

\textbf{Encapsulation ring} (Section~\ref{section_encapRing}) : The robot moves tangentially to the LOS vector $\zeta$ closely following the encirclement contour $O_g^{\mt{encap}}$ (Fig.~\ref{fig:zones}) with a step size $d_r$ such that it maintains safe distances from nearby robots. 

\subsubsection{Behavior Close to an Obstacle in the Secondary Zone}\label{section_obsSecZone}
\begin{figure}[b!]
    \centering
    \begin{subfigure}{0.35\textwidth}
    \includegraphics[width=\textwidth]{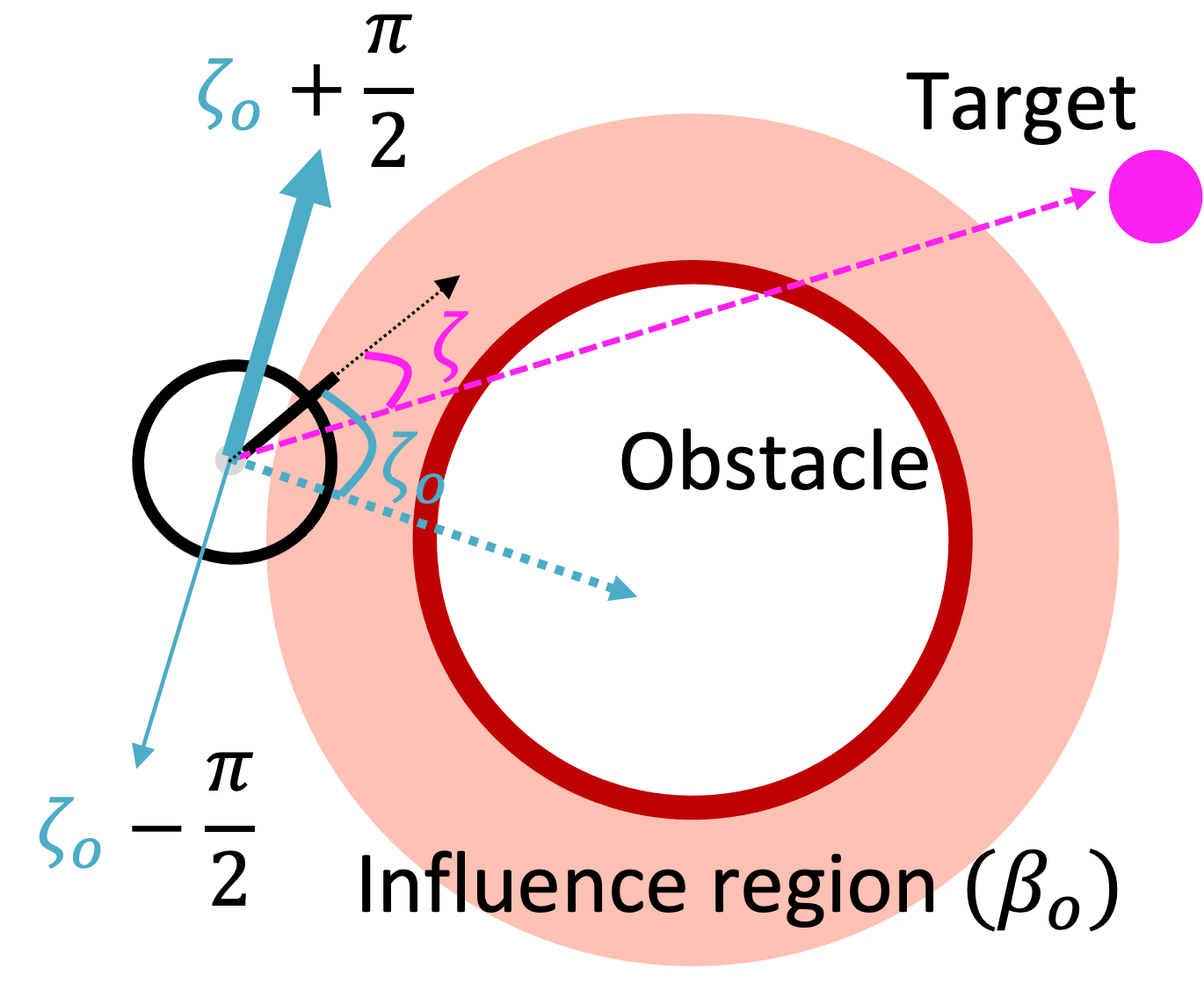}
     \caption{}
      \label{fig:obsFollowing}
    \end{subfigure}
    \begin{subfigure}{0.42\textwidth}
    \includegraphics[width=\textwidth]{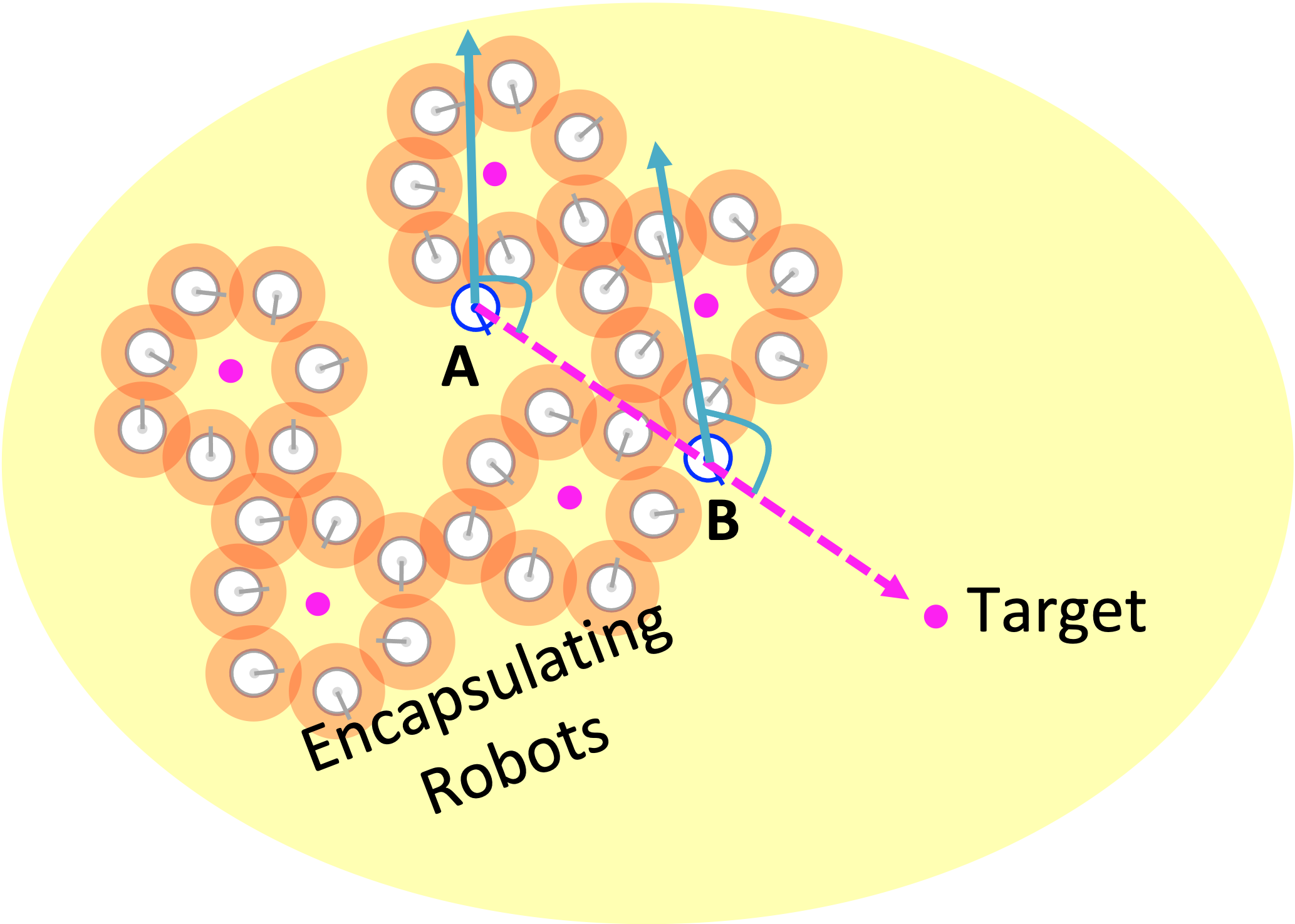}
     \caption{}
      \label{fig:chainedTargets}
    \end{subfigure}
    \caption{(a) A robot circumnavigates an obstacle in the tangential direction closest to the LOS vector, shown with a thicker teal arrow. (b) When targets are closer to each other than their minimum separation threshold $D_g$ (Section~\ref{section_minSep}), robots encapsulating them create an open obstacle. Robots at locations A and B both assess the obstacle as being behind them relative to the target's LOS vector $\zeta$, prompting them to advance toward the target. While this action is appropriate for the robot at B, it results in a livelock for the robot at A.
    }
\end{figure}
We want the robots to avoid obstacles while still moving towards the target. Thus, an efficient navigation strategy to avoid an obstacle is for a robot to deviate as little as possible from the target's LOS vector $\zeta$. To do so, we encode a ``wall-following'' behavior whenever a robot encounters an obstacle. A robot receiving obstacle signals ($Z_o \neq 0$) calculates $\nabla_{\mt{\mt{SG}}} F_o(\vb{x}^{\ell})$, where $\vb{x}^{\ell}$ is the $\ell^{\mt{th}}$ sensor's location in $\mathcal{B}$ with the highest obstacle signal intensity. 

Let $\zeta_o$ represent the direction toward the obstacle. To circumnavigate an obstacle, the robot moves tangentially with either $\theta_o^{\text{tan,-1}} = \zeta_o + \pi/2$ or $\theta_o^{\text{tan,+1}} = \zeta_o + 3\pi/2$ (clockwise or counterclockwise rotation, respectively). We choose the direction that minimizes the angular deviation from the target's LOS vector $\zeta$ (Fig.~\ref{fig:obsFollowing}). However, if the chosen direction deviates by more than $\pi/2$ from the robot's current heading, the robot sets $\theta_r = 0$, i.e. it continues with its current heading. This is to minimize alternating (avoiding livelocks) between clockwise ($\Theta_o^{\mt{tan},-1}$) and counterclockwise ($\Theta_o^{\mt{tan},+1}$), especially in scenarios where multiple nearby targets and obstacles influence the robot. The robot executes its obstacle circumnavigation strategy until the angle between $\zeta_o$ and $\zeta$ exceeds $\pi/2$, indicating that the obstacle is now behind the robot with respect to the closest target.

Following Assumption~\ref{assum_targetShutOff}, robots that have encapsulated a target also emit an obstacle signal $s_o$ and stop moving, essentially becoming stationary obstacles. This allows a robot to distinguish encapsulating robots from other dynamic robots nearby. Without this ability, the robot's lack of memory could result in livelocks around the ``ridge and trench'' formations created by encapsulating robots blocking the line of sight to a target; the robot would repeatedly advance toward the target, then retreat to avoid the encapsulating robots. 

\subsubsection{Behavior in the Encapsulation Ring}\label{section_encapRing}
A robot located at $\vb{x}_0 \in \mathcal{I}$ is in the encapsulation ring of the closest target $g$, if $\norm{\nabla f_g(r_g^{\mt{encap}})} < \norm{\nabla_{\mt{SG}} F_g(\vb{x}_0)} < \norm{\nabla f_g(r_g^{\mt{safe}})}$. To ensure that a robot remains in the encapsulation ring, we define an encircling orbit $O_g^{\mt{encap}} = r_g^{\mt{encap}} - d_r^{\mt{max}}$ as shown in Fig.~\ref{fig:zones}. We define the target attraction and avoidance angular ranges as: 
$
    \Theta_g^{\mt{LOS,-1}} = [\zeta, \textrm{ } \zeta+\pi/2] \textrm{ and }
    \Theta_g^{\mt{LOS,+1}} = [\zeta +3\pi/2, \textrm{ } \zeta] 
    $, 
    $
    \Theta_g^{\mt{avo,-1}} = [\zeta + \pi/2, \textrm{ } \zeta + \pi] \textrm{ and }
    \Theta_g^{\mt{avo,+1}} = [\zeta + \pi , \textrm{ } \zeta + 3\pi/2].
$
A robot's control then consists of the following behavior:\\
1. If $\norm{\nabla_{\mt{SG}}F_g(\vb{x}^k)} \leq \norm{\nabla f_g(O_g^{\mt{encap}})}$, the robot chooses a heading direction $\theta_r$ to move towards the target from the set $\Theta_g^{\mt{LOS},-1}$. If it cannot move in any of the directions in this set (i.e. $d_r = 0$), it chooses a heading from the set $ \Theta_g^{\mt{LOS},+1}$.\\
2. If $\norm{\nabla_{\mt{SG}}F_g(\vb{x}^k)} > \norm{\nabla f_g(O_g^{\mt{encap}})}$, the robot chooses the heading direction to avoid the target from the set $\Theta_g^{\mt{avo},+1}$. If it cannot move in any of the directions in this set (i.e. $d_r = 0$), it chooses a heading from the set $\Theta_g^{\mt{avo},-1}$.

The robot first tries to choose a heading in the clockwise orbit rotation ($\Theta_g^{\mt{LOS,-1}}$) if moving towards the target, and counterclockwise orbit rotation ($\Theta_g^{\mt{avo,+1}}$) if moving away from the target to ensure that robots moving towards and away from the target avoid oscillatory movements.

\section{Convergence Guarantees}\label{section_guarantees}
In this section, we assume noiseless sensors and derive constraints on various parameters that, if satisfied, guarantee that safety distances are maintained, the robots are never stuck in a deadlock or livelock, and all targets in the environment are eventually encapsulated. 

\subsection{Target Attraction}\label{section_boundTA}
Using the Lyapunov stability theorem \cite{Khalil, sinhmar2022decentralized}, we can show that a robot moves towards the target when the angular deviation ($\vartheta$) of its estimated LOS vector ($\zeta$) remains within $\pm\pi/2$ of the true LOS vector to the closest target. We determine constraints on the sensor placement, the number of sensors $p$, the safety distance $r_g^{\mt{safe}}$, and the source function $f_g$ that will guarantee motion toward the target. 

Let the maximum separation angle between any consecutive sensors on a robot be $\Phi$, the largest linear distance between them be $\Delta_H$ calculated as $2r_r\sin(\Phi/2)$, and the Lipschitz constant of $\nabla F_g$ in a ball of radius $\Delta_H$ centered at $\vb{x}^k$ be given by $L = \mt{max}(\norm{\nabla^2 F_g (\vb{x}^k)})$. Let the maximum gradient error be $\bm{\epsilon}$, then from Lemma~\ref{def_errSimplex} we have, $\norm{\nabla_{\mt{SG}} F_g(\vb{x}^k)-\nabla F_g(\vb{x}^k)}\leq L\frac{\Delta_H}{\sqrt{2}} \coloneq \norm{\bm{\epsilon}}$. We can show that the maximum possible angular deviation between the simplex gradient $\nabla_{\mt{SG}}F_g$ and true gradient $\nabla F_g$ is $\vartheta = \mt{sin}^{-1}\frac{\norm{\bm{\epsilon}}}{\norm{\nabla F_g(\vb{x}^k)}}$. The sufficient condition to ensure $\vartheta \in [-\pi/2, \pi/2]$ is $\norm{\bm{\epsilon}} < \norm{\nabla F_g(\vb{x}^k)}$, which is satisfied if the following two conditions are met:

\noindent\textbf{Condition 1: } the total number of sensors $p$ on a robot and the corresponding sensor placement is chosen such that $\frac{\Delta_H}{\sqrt{2}} \coloneq \sqrt{2}r_r\mt{sin}(\Phi/2) < 1$.

\noindent\textbf{Condition 2: } the Lipschitz constant of $\nabla f_g$ in the ball $B(\vb{x}^k,\Delta_H)$ satisfies $L < \norm{\nabla f_g(\vb{x}^k)}$. Intuitively, this means that the gradient of the source signal function $f_g$ varies smoothly throughout its domain, despite potentially high gradient magnitudes at specific locations. 

\begin{figure}
    \centering
    \includegraphics[width=\linewidth]{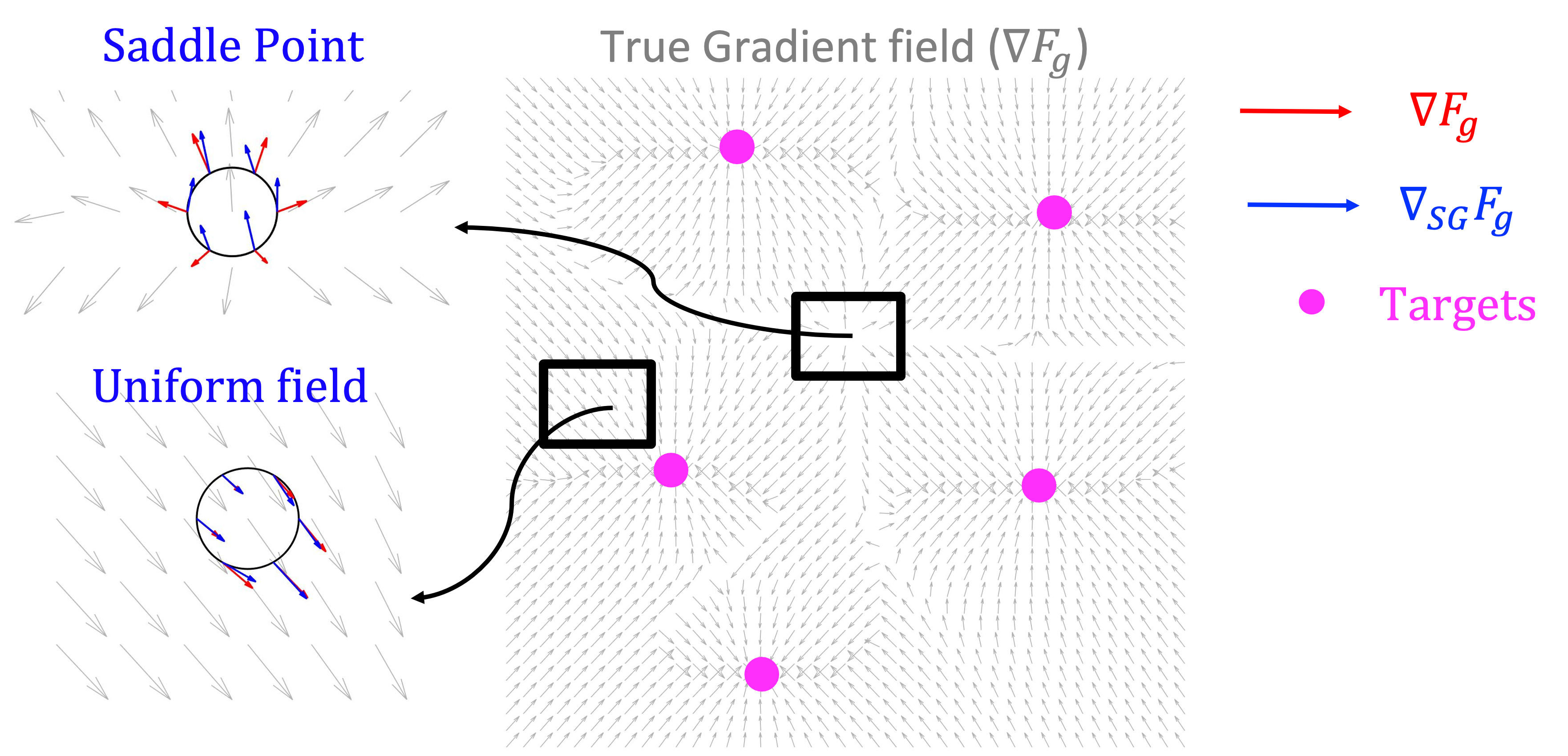}
    \caption{Gradient estimation deteriorates in critical region of $F_g$}
    \label{fig:gradEstimate}
\end{figure}

We examine these conditions for two possible cases: 
(I) regions with the influence of multiple targets ($F_g\neq f_g$), and (II) regions with the influence of a single target ($F_g = f_g$).
For (I), let $\vb{x}^k$ be such that there are no critical points of $F_g$ in the ball $B(\vb{x}^k,\Delta_H)$, and let the closest target be at $\vb{x}_g \in \mathcal{I}$ such that $d = \norm{\vb{x}^k - \vb{x}_g}$. Since $F_g$ inherits the continuity and differentiability properties of $f_g$ (Assumption~\ref{assum_fknown}),  $\norm{\nabla^2F_g(d)}< \norm{\nabla F_g(d)}$, implying $\norm{\bm{\epsilon}} \coloneq \norm{\nabla^2 F_g (d-\Delta_H)}\frac{\Delta_H}{\sqrt{2}} \leq \norm{\nabla F_g(d)}$ in the ball $B$, hence $\vartheta \in [-\pi/2, \pi/2]$.
Ensuring a constantly bounded angular error becomes challenging if one or more critical points exist inside the ball $B$ (Fig.~\ref{fig:gradEstimate}). By leveraging the embodiment of the robot ($r_r > 0$) and its distributed sensing, we set the robot heading in the direction indicated by the highest-intensity sensor reading ($\mt{max}(Z_g)$), so that the robot is guaranteed to get out of the local critical regions.
For (II), given the signal function $f_g$ and the number of sensors $p$, we set constraints on $r_g^{\mt{safe}}$ such that the following inequality is satisfied,
    \begin{align}
        \norm{\nabla^2 f_g(r_g^{\mt{safe}})}\frac{\Delta_H}{\sqrt{2}} < \norm{\nabla f_g(r_g^{\mt{safe}})}
        \label{eq_limit_rgsafe}
    \end{align}
Fig.~\ref{fig:gradEstimate} shows how, for a given number of sensors, gradient estimate deteriorates in the critical regions of scalar field $F_g$.\\

\subsection{Guaranteed Encirclement Behavior} \label{section_guaranteeEncap}
Next, we establish additional constraints on the signal function $f_g$ and the minimum distance between any two targets, so that a robot can detect when it is within a target's encapsulation ring and start the encirclement behavior. Consider an environment with $m \geq 1$ targets and a robot in a configuration such that target $g_1$ located at $\vb{x}_{g_1} \in \mathcal{I}$ is closest to the robot at any time $T$. Let the $k^{\mt{th}}$ sensor receive maximum target intensity ($z_g^k = \mt{max}(Z_g)$) and $r_{g_1}^{\mt{safe}}$, $r_{g_1}^{\mt{encap}}$ be the inner and outer radius of the encapsulation ring of the closest target $g_1$, respectively (bounds on encapsulation ring radii provided in~\cite{sinhmar2022decentralized}). For a robot to enter the encapsulation ring without breaching the safety violation zone, the following conditions must be satisfied:

\noindent\textbf{Condition 3:} The gradient of the aggregate signal strength $F_g$ at the inner radius, $\norm{\nabla F_g(r_{g_1}^{\mt{safe}})}$, must exceed the gradient of single target influence at the outer radius $\norm{\nabla f_g(r_{g_1}^{\mt{encap}})}$, that is $\norm{\nabla F_g(r_{g_1}^{\mt{safe}})} > \norm{\nabla f_g(r_{g_1}^{\mt{encap}})}$ 

\noindent\textbf{Condition 4:} Conversely, the gradient of the aggregate signal strength $F_g$ at the outer radius, $\norm{\nabla F_g(r_{g_1}^{\mt{encap}})}$, must be less than the gradient of single target influence at the inner radius $\norm{\nabla f_g(r_{g_1}^{\mt{safe}})}$, that is $\norm{\nabla F_g(r_{g_1}^{\mt{encap}})} < \norm{\nabla f_g(r_{g_1}^{\mt{safe}})}$

These are the necessary conditions to ensure that the encirclement behavior is triggered only if a robot is in the encapsulation ring as detailed in Section~\ref{section_encapRing}. Furthermore, for successful encapsulation of a target $g$, it is necessary that within the encapsulation ring, the target's influence dominates, i.e. $F_g \approx f_g$ for $\norm{\vb{x}^k-\vb{x}_g} \leq r_g^{\text{encap}}$. Therefore, we require a minimum separation $D_g \in \mathbb{R}^+$ between any two targets in the environment. The lower bound on $D_g$ is a function of both the number of sensors and their placement, and $f_g$.  We elaborate on two specific scenarios to compute $D_g$. These cases depict the extreme target configurations with all other scenarios being variations of these extremes.\\
\noindent\textbf{Case I:} All other $m-1$ targets are located on the opposite side of the robot with the next closest target located at $\vb{x}_{g_2}$ such that $D_g = \norm{\vb{x}_{g_1}-\vb{x}_{g_2}}$. In this case, $\norm{\nabla F_g(\vb{x}^k)} < \norm{\nabla f_g(\vb{x}^k)}$. If the $k^{\mt{th}}$ sensor is located at $\vb{x}^k \in \mathcal{I}$ such that it is at the inner radius $r_{g_1}^{\mt{safe}}$ then Condition 3 is satisfied if, 
\begin{align}\label{eq_infDg}
    \inf \norm{\nabla F_g(\vb{x}^k)} \! \coloneq \!\norm{\nabla f_g(r_{g_1}^{\mt{safe}})} \!-\!(m\!-\!1)\norm{\nabla f_g(D_g-r_{g_1}^{\mt{safe}})} \! &> \! \norm{\nabla f_g(r_{g_1}^{\mt{encap}})} 
\end{align}
\noindent\textbf{Case II:} All the other $m-1$ targets are located on the same side of the robot with the second closest target located at $\vb{x}_{g_2}$ such that $D_g = \norm{\vb{x}_{g_1}-\vb{x}_{g_2}}$. In this case, $\norm{\nabla F_g(\vb{x}^k)} > \norm{\nabla f_g(\vb{x}^k)}$. If the $k^{\mt{th}}$ sensor is located at $\vb{x}^k \in \mathcal{I}$ at the outer radius $r_{g_1}^{\mt{encap}}$ then Condition 4 is satisfied if,
\begin{align}\label{eq_supDg}
    \sup \norm{\nabla F_g(\vb{x}^k)} \! \coloneq \! \norm{\nabla f_g(r_{g_1}^{\mt{encap}})}\! +\!  (m\! -\! 1)\norm{\nabla f_g(D_g-r_{g_1}^{\mt{encap}})} \! &<\! \norm{\nabla f_g(r_{g_1}^{\mt{safe}})} 
\end{align}
Note that Condition 4 is satisfied for all $D_g$ for Case I, and Condition 3 is satisfied for all $D_g$ for Case II. 

\noindent \textbf{Example:}
One example of a target source signal function that satisfies Conditions 1 and 2 is the radially decreasing function $f_g = C/d^2$ where $C \in \mathbb{R}^+$ and $d$ is the distance from the source.
For a circular robot with $p$ sensors placed symmetrically on its periphery, the constraints on the safety distance $r_g^{\mt{safe}}$ and the minimum separation between targets $D_g$ are given below such that Condition 3 and Condition 4 are satisfied. 
From Eq.~\eqref{eq_limit_rgsafe} we have, $$\frac{6C}{(r_g^{\mt{safe}})^4}\sqrt{2}r_r \mt{sin}(\pi/p) < \frac{2C}{(r_g^{\mt{safe}})^3} \implies r_g^{\mt{safe}} \geq 3\sqrt{2}r_r\mt{sin}(\pi/p)$$
From Eq.~\eqref{eq_infDg} and Eq.~\eqref{eq_supDg}, $$D_g > \mt{max}\Bigg\{r_g^{\mt{safe}}\bigg[1+\frac{m-1}{\big[1-\big(\frac{r_g^{\mt{safe}}}{r_g^{\mt{encap}}}\big)^3\big]}\bigg]^{1/3}, r_g^{\mt{encap}}\bigg[1+\frac{m-1}{\big[\big(\frac{r_g^{\mt{encap}}}{r_g^{\mt{safe}}}-1\big)^3\big]}\bigg]^{1/3}\Bigg\}$$

\subsection{Absence of Livelocks: Minimum Separation Between Obstacles}\label{section_minSep}
To ensure no livelocks, the minimum separation $D_o$ required between any two obstacles for a robot to navigate between them is given by
\begin{align}\label{eq_boundObsSep}
    D_o &> 2r_{\mt{o}} + 2(r_r + r_{o}^{\mt{safe} \mid \mt{sensor}}) + d_r^{\mt{max}}\\
    r_{o}^{\mt{safe} \mid \mt{sensor}} &= \sqrt{(r_o^\mt{safe})^2 + r_r^2 - 2r_rr_o^\mt{safe}\mt{cos}(\pi/p)} \nonumber
\end{align}
where $r_o$ is the obstacle's circumscribed radius, $r_{o}^{\mt{safe} \mid \mt{sensor}}$ is the safe distance $r_o^{\mt{safe}}$ with respect to a robot's sensor. $D_o$ is a function of the number of sensors $p$: a robot with more sensors can navigate through more closely spaced obstacles. 

Additionally, since encapsulating robots emit obstacle signals, we need to ensure that multiple sets of encapsulated targets do not form a non-convex open region in the environment, as shown in Fig.~\ref{fig:chainedTargets}. Therefore, we put additional constraints on the minimum separation between any two targets,
\begin{align}\label{eq_D_gBound}
    D_g > 2r_g^{\mt{encap}} + 2(r_r + r_g^{\mt{safe}\mid\mt{sensor}}) + d_r^{\mt{max}}
\end{align}

Satisfaction of Conditions 1-4, Eq.~\eqref{eq_limit_rgsafe}-\eqref{eq_D_gBound} and  Lyapunov stability guarantee that the robot moves toward the closest target and eventually reaches its encapsulation ring. When the robot encounters an obstacle, it moves away 
to successfully circumnavigate the obstacle while maintaining its overall motion toward the target, ensuring a control policy that combines obstacle avoidance with target attraction. 

\section{Simulations}
We examine how the number of sensors ($p$), sensor noise, occlusions, deviations from the determined bounds on $f_g$, and minimum target separation $D_g$ affect the swarm's emergent behavior. 
The environment comprises 3 targets, 6 circular obstacles, and 18 robots. The total simulation time is capped at 1500 time steps. We run 50 simulations for each data point with the same initial conditions to study the effect of the considered parameter on the emergent behavior by keeping other conditions constant. Stochasticity within each simulation originates from the random selection of $\theta_r$. All robots in the swarm move at different rotational and translational speeds.\\

\begin{figure}[!t]
  \centering
  \includegraphics[width=0.4\linewidth]{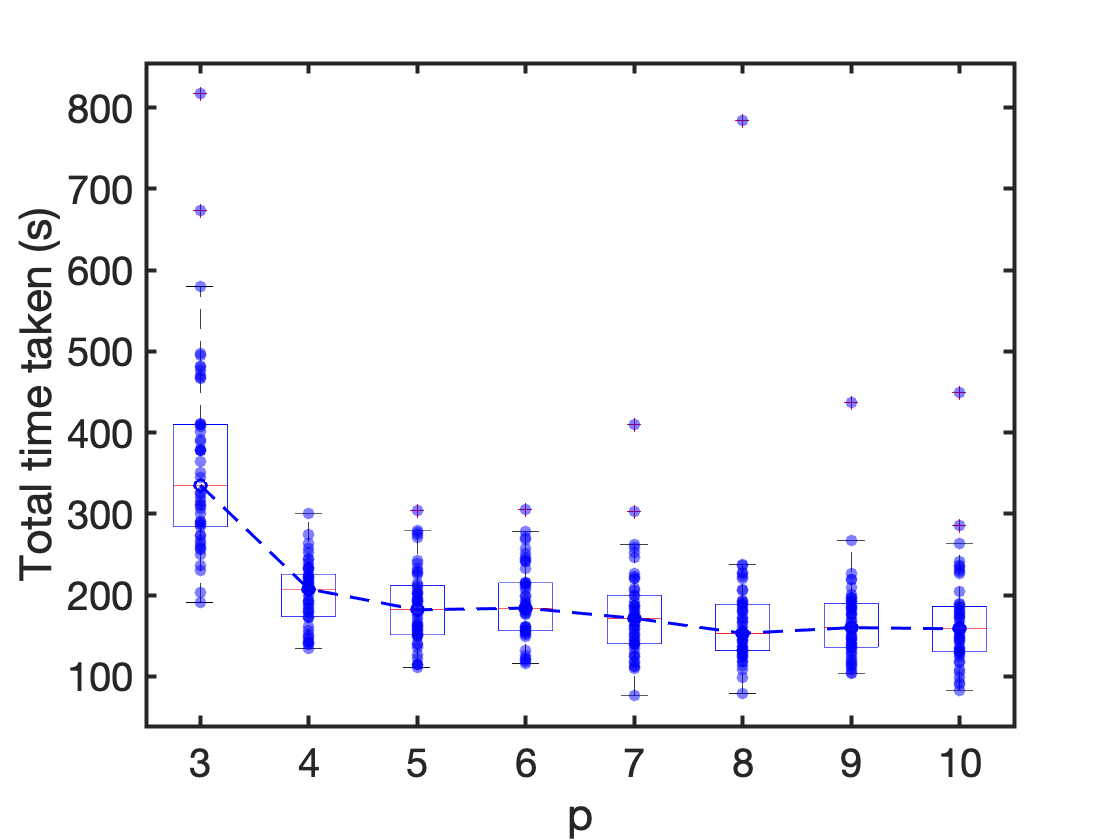}
  \caption{The total time taken for task completion as a function of $p$.}
  \label{fig:p_totalTime}
\end{figure}

\noindent\textbf{Effect of the total number of sensors $(p)$:}
Fig. \ref{fig:p_totalTime} shows how increasing the number of sensors reduces the time required to encapsulate all targets while ensuring safety. Here $n_g = 6$ and the control is synchronous. This is because when $p$ increases, the error in computing the simplex gradient decreases, resulting in a better \textit{impression} of the local neighborhood. 
\\

\begin{figure}
    \centering
    \begin{subfigure}{0.32\textwidth}
    \includegraphics[width=\textwidth]{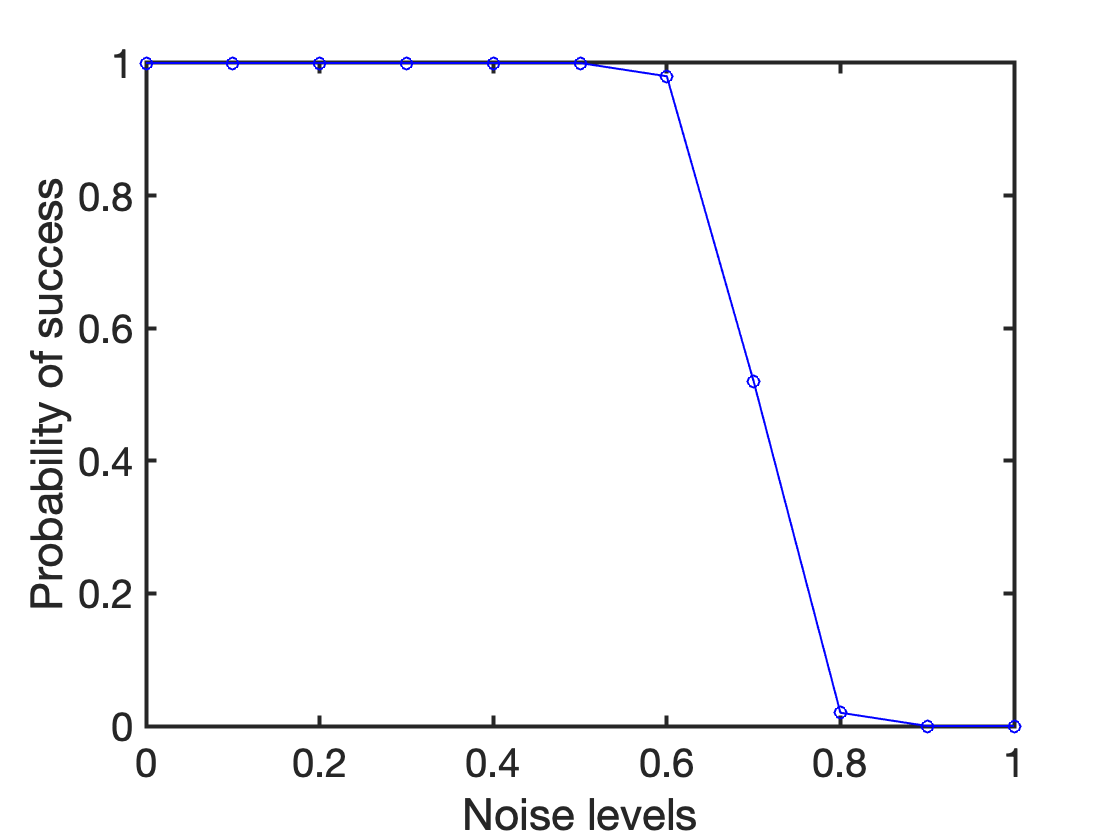}
     \caption{}
      \label{fig:noise_prob}
    \end{subfigure}
    \begin{subfigure}{0.32\textwidth}
    \includegraphics[width=\textwidth]{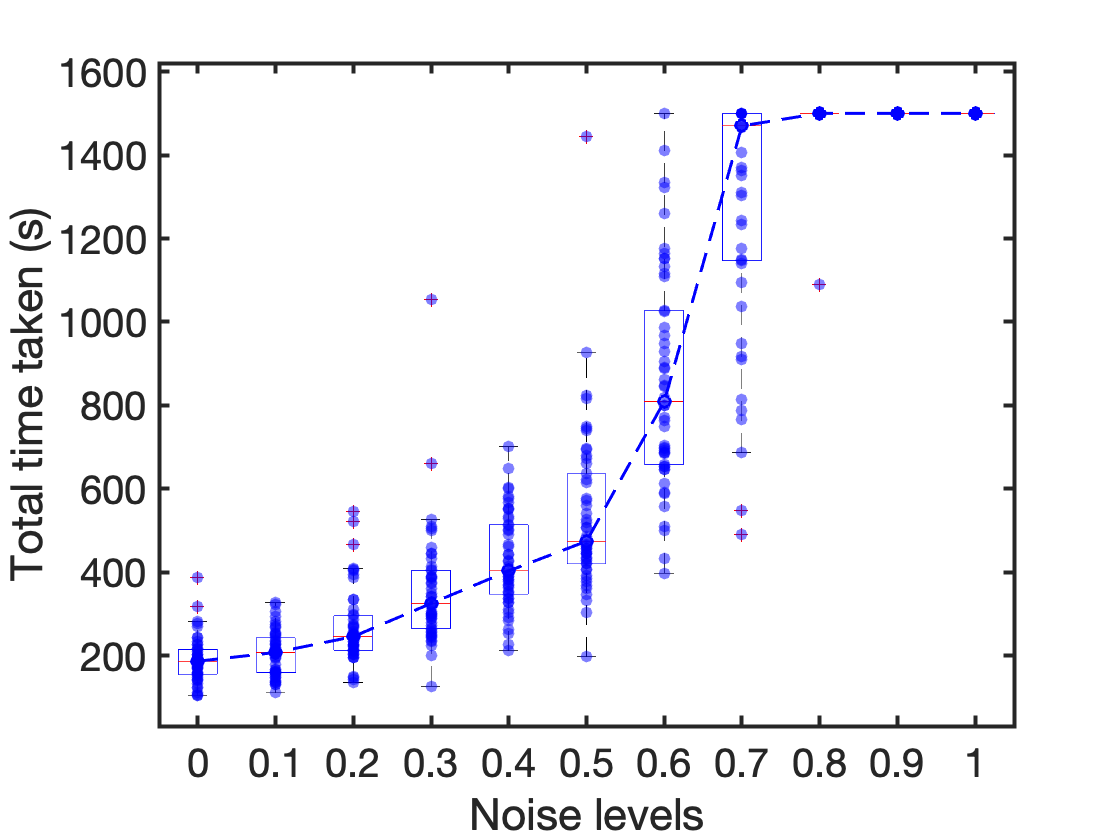}
     \caption{}
      \label{fig:noise_totalTime}
    \end{subfigure}
    \begin{subfigure}{0.32\textwidth}
    \includegraphics[width=\textwidth]{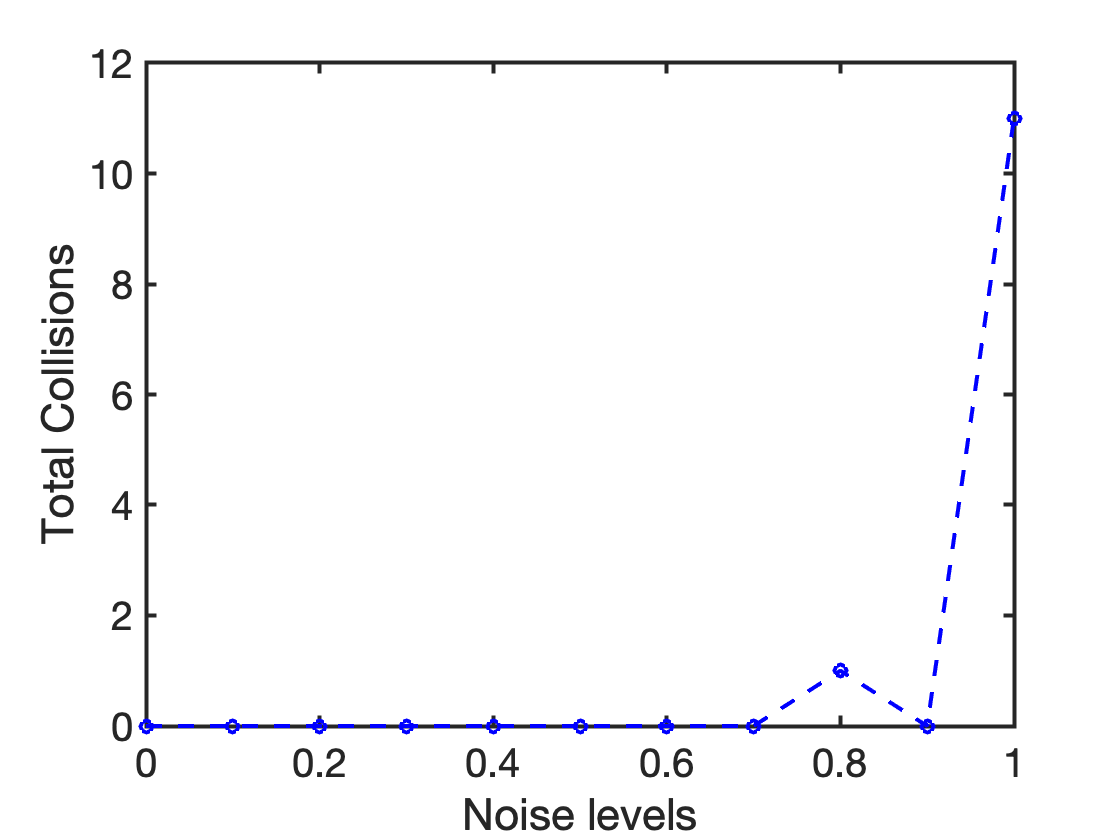}
     \caption{}
      \label{fig:noise_coll}
    \end{subfigure}
    \caption{For $p = 5$ (a) probability of success as a function of noise levels (b) time taken for task completion (c) number of $r_{g}^{\mt{safe}}$ violations across all simulations.} 
    \vspace{0.7em}
\end{figure}

\noindent\textbf{Effect of noise:} 
We add Gaussian noise to each sensor reading, $z_g^k = (1 - n_g^k)\sum_{j\in N_g^k} f_g(d^{k}_g),$ $\, n_g^k\sim \mathcal{N}(0,\sigma^2)$ and $n^k_g \leq 1$. 
Fig.~\ref{fig:noise_totalTime} shows the total time taken by the swarm for task completion with $p = 5$. As the noise increases, the deviation of the simplex gradient from the true gradient increases, leading to inaccurate estimations of the closest target's encapsulation ring. Consequently, the success rate decreases considerably for noise levels exceeding 70\% and drops to zero when the noise threshold surpasses 85\%. Unlike
~\cite{sinhmar2023guaranteed}, where task parameters (e.g., $r_g^{\mt{encap}}$,$n_g$, $r_g^{\mt{safe}}$) had to be adjusted to ensure successful encapsulation, here we keep all parameters fixed and show that 
the use of simplex gradients, which leverage distributed sensing on a robot, enhances robustness to noise.
Contrary to~\cite{sinhmar2022decentralized}, the addition of noise to sensor readings for robot sources did not result in noise filtering, and we observed safety violations. This difference stems from our relaxation of the robot's step size upper bound (Section~\ref{section_ca}). 
\\

\begin{figure}[!t]
    \centering
    \begin{subfigure}{0.32\textwidth}
    \includegraphics[width=\textwidth]{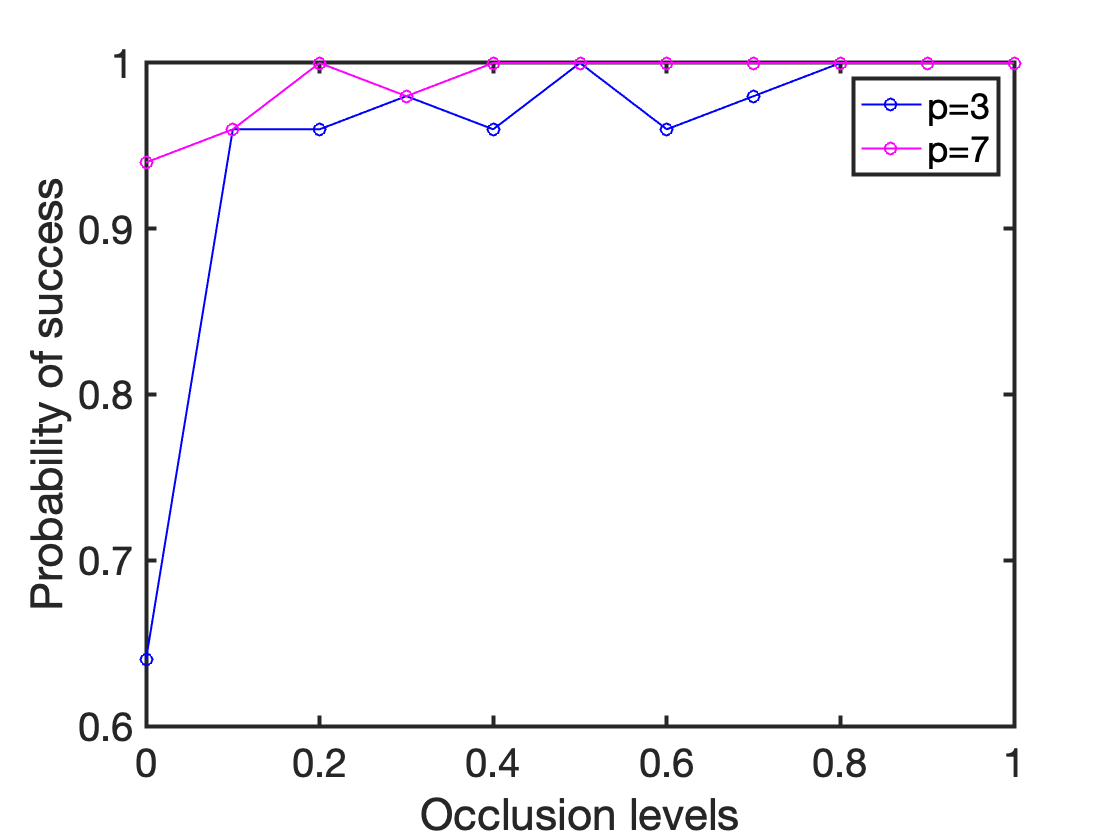}
     \caption{}
      \label{fig:occ_prob}
    \end{subfigure}
    \begin{subfigure}{0.32\textwidth}
    \includegraphics[width=\textwidth]{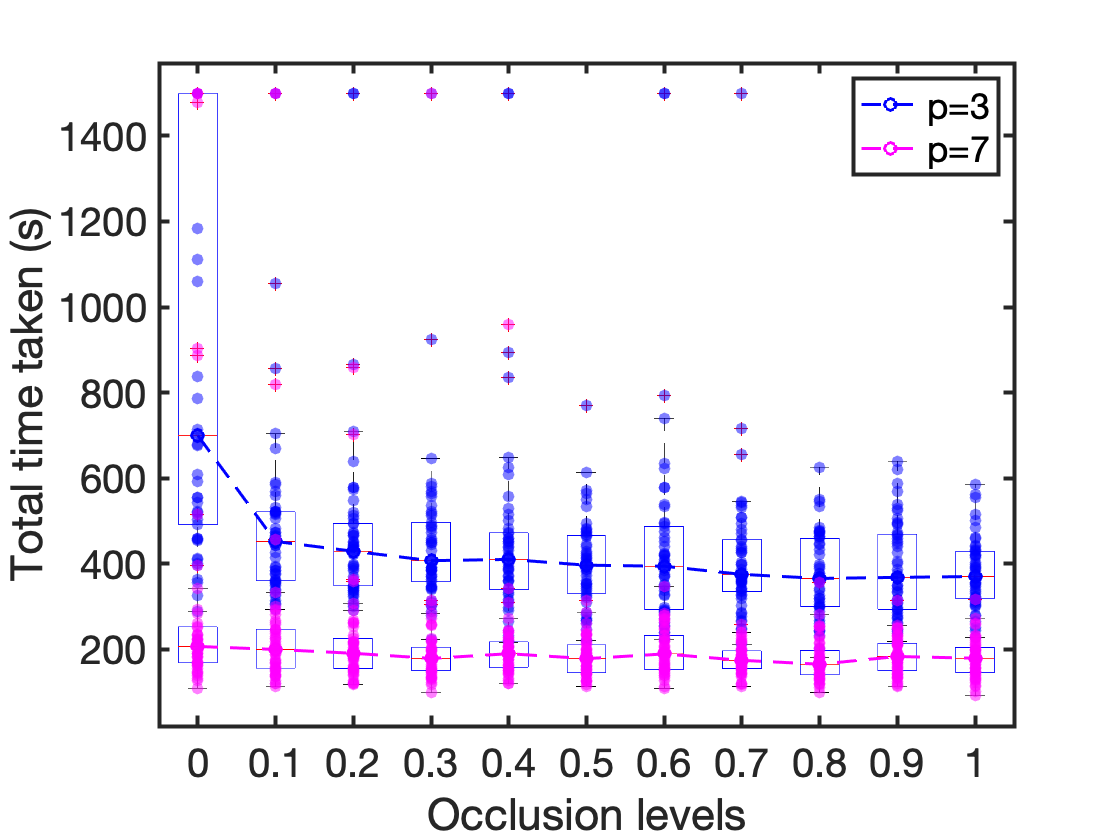}
     \caption{}
      \label{fig:occ_totalTime}
    \end{subfigure}
    \begin{subfigure}{0.32\textwidth}
    \includegraphics[width=\textwidth]{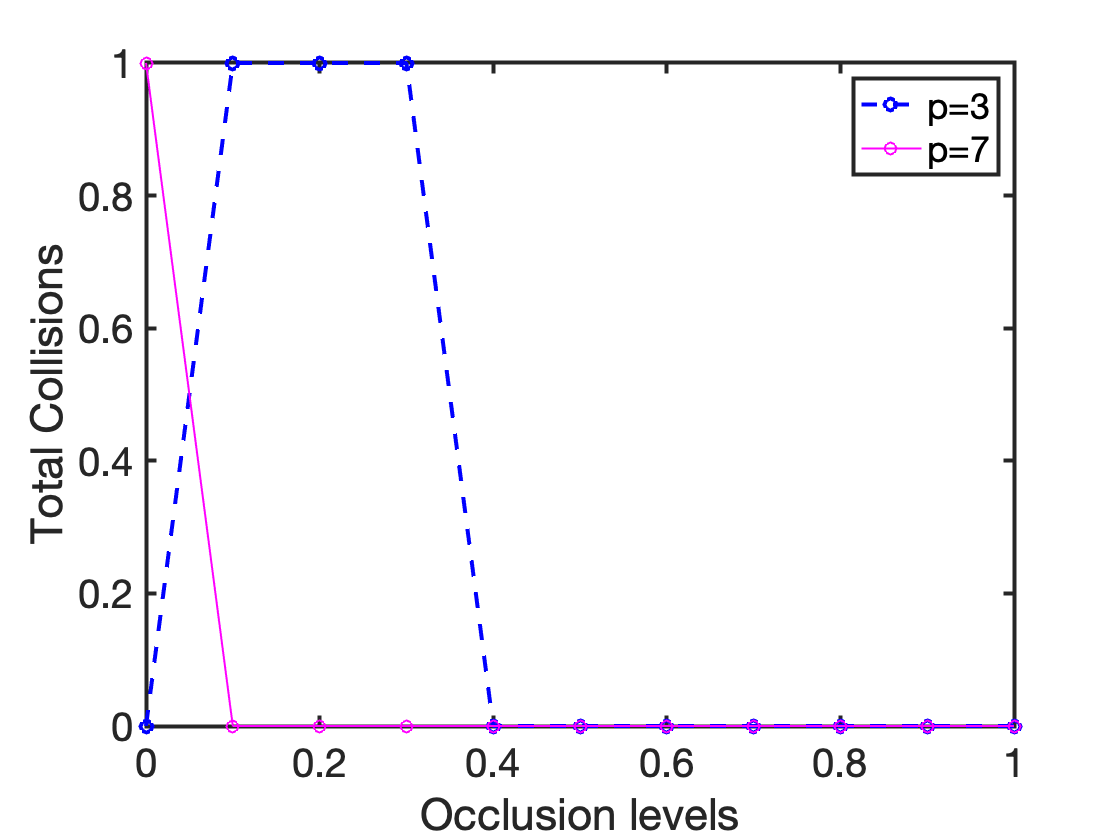}
     \caption{}
      \label{fig:occ_coll}
    \end{subfigure}
    \caption{(a) probability of success as a function of occlusion levels (b) total time taken for task completion. (c)  number of safety violations across all simulations}
    \vspace{.3em}
\end{figure}

\noindent\textbf{Effect of occlusions:}
We adjusted the sensor measurement $z_s^k$ by applying a coefficient $\eta\in[0,1]$ to quantify the obstruction level between the source and the sensor. This adjustment accounts for scenarios where obstacles or other robots obstruct the line of sight from the source to the $k^{\text{th}}$ sensor. In this scale, 0 indicates complete obstruction, and 1 indicates full visibility. 
We investigate the effects of varying the total number of sensors ($p$) on a robot, considering different levels of occlusion. Figure~\ref{fig:occ_prob} illustrates that the success probability rises from 65\% to 100\% for $\eta \geq 0.75$ with $p = 3$. Increasing the sensor count to $7$ boosts success rates, with a minimum of 94\% at $\eta = 0$ and reaching 100\% for $\eta \geq 0.45$, indicating improved robustness with higher sensor resolution. 
\\

\noindent\textbf{Supplementary Video \cite{supplvideo}:} We include simulations that show the effect of constraint violations, asynchronous execution, and a demonstration of the algorithm's scalability through a large-scale simulation involving 100 robots, 15 targets, and 10 obstacles.


\section{Conclusion}
In this paper, we show how robots equipped with omnidirectional sensors and an isotropic signal emitter can successfully encapsulate multiple diffusive targets while maintaining safety distances from other robots and obstacles. Our decentralized controller is agnostic to the number of robots, targets, and obstacles in the environment. We determined bounds and constraints on task, control, and robot parameters for guaranteed convergence to all sources with user-specified safety constraints. Additionally, our algorithm can be adapted for robots with non-isotropic sensors, provided the measurement error bounds are known. In future research, we will explore relaxing the assumption that a target source stops signal emission upon encapsulation, which currently directs non-encapsulating robots to search for other targets. One strategy could be to extend the influence region of encapsulating robots, prompting dynamic robots to avoid these regions and explore remaining active target sources.

\bibliographystyle{styles/bibtex/splncs03_unsrt}
\bibliography{ref}
\end{document}